\documentclass[journal]{IEEEtran}
\usepackage{cite}
\usepackage{amsmath,amssymb,amsfonts}
\usepackage{algorithmic}
\usepackage{graphicx} 
\usepackage{commath}
\usepackage{amsmath}
\usepackage{graphicx}
\usepackage{multicol}
\usepackage{multirow}
\usepackage{caption}
\usepackage{subcaption}
\usepackage{lscape}
\usepackage{lmodern,adjustbox,booktabs}
\usepackage{textcomp}
\ifCLASSINFOpdf
\else
\fi
\begin{document}
%
\title{The Deep Radial Basis Function Data Descriptor (D-RBFDD) Network: A One-Class Neural Network for Anomaly Detection}
%
%
%

\author{\IEEEauthorblockN{Mehran H. Z. Bazargani, Arjun Pakrashi, Brian Mac\ Namee}\\
\IEEEauthorblockA{School of Computer Science\\Insight Centre for Data Analytics \\
University College Dublin (UCD)\\
Dublin, Ireland\\
mehran.hosseinzadehbazarga@ucdconnect.ie, \{arjun.pakrashi,brian.macnamee\}@ucd.ie}


\thanks{}}


%
%

\markboth{IEEE Transactions on Neural Networks and Learning Systems: Special Issue on Deep Learning for Anomaly Detection}%
{Shell \MakeLowercase{\textit{et al.}}: IEEE Transactions on Neural Networks and Learning Systems: Special Issue on Deep Learning for Anomaly Detection}
%



\maketitle

\begin{abstract}
Anomaly detection is a challenging problem in machine learning, and is even more so when dealing with instances that are captured in low-level, raw data representations without a well-behaved set of engineered features. The Radial Basis Function Data Descriptor (RBFDD) network is an effective solution for anomaly detection, however, it is a shallow model that does not  deal effectively with raw data representations. This paper investigates approaches to modifying the RBFDD network to transform it into a deep one-class classifier suitable for anomaly detection problems with low-level raw data representations. We show that approaches based on transfer learning are not effective and our results suggest that this is because the latent representations learned by generic classification models are not suitable for anomaly detection. Instead we show that an approach that adds multiple convolutional layers before the RBF layer, to form a Deep Radial Basis Function Data Descriptor (D-RBFDD) network, is very effective. This is shown in a set of evaluation experiments using multiple anomaly detection scenarios created from publicly available image classification datasets, and a real-world anomaly detection dataset in which different types of arrhythmia are detected in electrocardiogram (ECG) data. Our experiments show that the D-RBFDD network out-performs state-of-the-art anomaly detection methods including the Deep Support Vector Data Descriptor (Deep SVDD), One-Class SVM, and Isolation Forest on the image datasets, and produces competitive results for the ECG dataset.
\end{abstract}

\begin{IEEEkeywords}
Anomaly detection, Deep Learning, RBF networks, Artificial Neural Networks.
\end{IEEEkeywords}

%
\IEEEpeerreviewmaketitle

\section{Introduction}
\label{sec:introduction}
Chandola \& Kumar \cite{Anomaly} define \textit{anomaly detection} as ``\textit{the problem of finding patterns in data that do not conform to expected behavior}''. Anomaly detection is made especially challenging by the lack of access to anomalous patterns during training. \textit{One-class classification} \cite{DBLP:journals/corr/KhanM13}, in which a model is trained to recognize normal data and flag an anomaly when something fails to be recognized as normal, is a common approach to building anomaly detectors. Like any other machine learning task, however, training a model can be challenging when dealing with raw data as opposed to a set of well-behaved engineered features---for example, when working with image data \cite{zhou2017anomaly}, audio data \cite{rushe2019anomaly}, or streaming data \cite{ahmad2017unsupervised}. Shallow models particularly suffer from this issue, and it is common practice to employ deep learning in these scenarios.

In this paper we propose a deep anomaly detection model that can be trained in a fully end-to-end fashion, and is suitable for use with raw, high-dimensional data sources such as images and timeseries data from sensors. Our proposed model, the \emph{Deep Radial Basis Function Data Descriptor} (D-RBFDD) network, is based on our previous work on the Radial Basis Function Data Descriptor (RBFDD) network \cite{RBFDD}. RBFDD networks (which in turn are based on Radial Basis Function (RBF) networks \cite{Ethem,Bishop}) are  effective, efficient anomaly detectors that learn a compact set of Gaussian kernels to cover the normal region of input space, and recognize anomalies as instances that sit outside this region. The positions of these learned kernels, and the magnitude of the weights connecting each of them to the output layer, also facilitate straight-forward post-hoc explanation of outputs \cite{jin2003extracting, xi2018interpretable, augusteijn2000constructing}.

Although RBFDD networks are effective, they are shallow neural networks with a single hidden layer and do not perform well when trained on raw data \cite{RBFDD}. We identify three ways to adapt RBFDD networks to work with raw input data: (1) based on transfer learning using the latent representation from a generic pre-trained network as input to the RBFDD network, (2) extending the first approach to include fine-tuning the pre-trained network as part of training the RBFDD, and (3) adding multiple computational layers before the RBFDD network that are fully trained as part of training the network in an end-to-end fashion.

In this paper we explore the effectiveness of these three approaches, and show that the final approach that trains the entire network from random initialization---referred to as Deep RBFDD (D-RBFDD)---out-performs the other deepening approaches. This addresses a fundamental question of whether the latent representations learned by large models trained for multi-class classification are suitable as input for anomaly detection models, or whether they are too entangled with the multi-class classification problem. Our evaluation experiments---using anomaly detection scenarios constructed from well-known image classification datasets and a dataset from a real-world electrocardiogram (ECG) anomaly detection task---suggest that they are not suitable. Our experimental results also show that the D-RBFDD approach out-performs existing state-of-the-art anomaly detection approaches (including the shallow RBFDD approach) on the image datasets, while producing competitive results on the ECG dataset. D-RBFDD is therefore an effective anomaly detector trained in an end-to-end fashion, that has the advantages that come with an approach based on RBF networks: it is efficient, it lends itself to easy interpretation, and the local learning approach can adapt to dynamic definitions of normality and can accommodate changing distributions (i.e., concept drift when learning from streams). 

The remainder of this article is structured as follows. In Section \ref{Related Work} we describe common approaches to anomaly detection including deep learning methods. In Section \ref{RBFDD networks}, we describe the RBFDD network approach and illustrate different methods for deepening it. The setup of our experiments is described in Section \ref{Experiment Setup}, and the results are discussed in Section \ref{Results and Discussions}. Finally, Section \ref{Conclusion and Futurework} concludes the paper and discusses some directions for future work.

\section{Related Work}\label{Related Work}
Machine learning approaches to anomaly detection are dominated by a family of algorithms that adapt the Support Vector Machine (SVM) \cite{Ethem} algorithm to work with only examples of a single class: One-Class SVM (OCSVM) \cite{OCSVM}. In fact Khan \& Madden \cite{DBLP:journals/corr/KhanM13} go so far as to say that one-class classification approaches should be divided into two categories: OCSVMs and everything else! 

Much like the standard SVM approach, OCSVM models separate normal data points from the origin in the feature space using a hyper-plane, while maximizing the distance between the origin and this hyper-plane. Although any kernel function can be used, Gaussian kernels work particularly well for OCSVM models \cite{DBLP:journals/corr/KhanM13}. The main issue with OCSVM models is that they do not scale well. For large datasets the computational and storage requirements of OCSVMs grow polynomially with dataset size \cite{Awad2015}. Variations of the OCSVM approach include the Support Vector Machine Data Description (SVDD) \cite{Tax:2004:SVD:960091.960109}, which uses hyper-spheres rather than hyper-planes to achieve separation. The goal is to find a spherically shaped boundary, encompassing the normal data.

On the non-OCSVM side, Isolation Forests (iForests) \cite{iForest} and Auto-Encoder neural networks (AENs) \cite{GoodFellowDeepLearning} (and their many variations) are effective anomaly detectors. An iForest model isolates individual data points in a training set by constructing a decision tree that splits the input space randomly and repeatedly. The intuition behind this approach is that fewer splits should be required to isolate anomalous instances than normal ones. An auto-encoder network (AEN) is a neural network that learns a generative model of input data by transforming it into a representation with reduced dimensionality, and then reconstructing the original input data from this representation. If an AEN is trained to reconstruct only \emph{normal} data instances, it can detect anomalies by flagging test instances for which the reconstruction error is very high. Deep auto-encoders have been shown to be effective on problems with raw data inputs \cite{an2015variational,zhao2017spatio}. 

Deep learning approaches for anomaly detection can be categorized as either \textit{mixed approaches} or \textit{end-to-end approaches}. In \emph{mixed approaches} a deep model is trained in an unsupervised way to work as a feature extractor that produces the data for a, typically shallow, anomaly detector, such as a OCSVM. The deep models used to learn features tend to be reconstruction-based models such as Deep Belief Networks (DBNs) or deep AENs \cite{RecognitionOfGeochemicalAnomaliesUsingADeepAutoencoderNetwork, OnAccurateandReliableAnomalyDetectionforGasTurbineCombustorDeepLearningApproach, AdeepLearningApproachforDetectingMaliciousJavaScriptCode}. For example, in \cite{erfani2016high} an unsupervised DBN is trained to extract generic underlying features, and a OCSVM is trained using these features. The mixture of a DBN and a OCSVM is shown to out-perform a standalone OCSVM. 

Similarly, in \cite{marir2018distributed}, in order to detect anomalous behavior in large-scale network traffic data, a DBN model is trained as an unsupervised dimensionality reduction step, whose output features are then fed into a multi-layered ensemble Support Vector Machine (SVM). In \cite{xu2017detecting}, a fully unsupervised model is proposed for detecting anomalous frames in video. For every frame of the video, appearance and motion features are extracted, and fed into two separate Stacked Denoising Auto-Encoders (SDAEs). A fusion of these two types of features are fed into a third SDAE. The features obtained in the bottle-neck layers of the three SDAEs are then fed into three OCSVMs, each of which produces an anomaly score. The three anomaly scores are combined to make the final decision for an input video frame. 

The main issue with mixed approaches is that the deep feature extractor is not trained for an anomaly detection objective, but on a different objective such as minimizing the reconstruction error. As a result, the features learned may not be useful inputs for the anomaly detection model. \textit{End-to-end approaches} address this issue, and aim to make the features extracted in latent representations more appropriate for the anomaly detection task by defining a one-class loss function. The loss function is then used to train an entire network in an end-to-end fashion, guiding the network to produce features that are appropriate for anomaly detection. For example, in \cite{zong2018deep} the Deep Auto-encoding Gaussian Mixture Model (DAGMM) is proposed as an end-to-end approach to anomaly detection. An AEN is used to reduce the dimensionality of the data, while the reconstruction error and the low-dimensional representation from the bottle-neck of the AEN are fed into a Gaussian Mixture Model (GMM) \cite{Ethem}. Similarly, in \cite{chalapathy2018anomaly} a One-Class Neural Network (OCNN) for anomaly detection is proposed. OCNN combines the rich feature extraction property of deep neural networks with a proposed OCSVM-like cost function. At first, a deep AEN is trained to produce good representative features of the input data. Next, the encoder portion of this pre-trained AEN is fed into a simple one-layer neural network, the ultimate output of which will be used to compute the cost. The weights of both the encoder and the one-layer neural network will be trained simultaneously, while minimizing the cost function. By combining the capability of deep neural networks to extract progressively rich features from the data with the proposed cost function, the aim is to obtain the hyper-plane that separates the normal data from the origin.

AnoGAN \cite{schlegl2017unsupervised} is a deep model for anomaly detection based on Generative Adversarial Networks (GANs) \cite{GoodFellowDeepLearning}. The generator network is trained to learn the distribution of the training data. Given a test instance it searches for a point in the latent space of the generator that would generate a sample that is closest to the test point. If an accurate distribution of the normal data has been learned, for a normal query, $x$, there should be a representation, $z$, in the latent space that the generator could use to generate a new data point, $G(z)$, that is similar to the normal query $x$. For an anomalous query a good representation, $z$, should not be found and, as a result, the generated data, $G(z)$, will not be similar to the query.

Finally, inspired by the Support Vector Data Descriptor (SVDD) model \cite{Tax:2004:SVD:960091.960109}, Deep-SVDD  \cite{pmlr-v80-ruff18a} is another deep one-class neural network designed for anomaly detection. While the neural network is trained, the volume of a hypersphere that envelopes the data in the latent space is minimized. Thus, the neural network is trained to map the input data into a hypersphere with minimum volume. There are two versions of Deep SVDD: (1) Soft-boundary Deep SVDD which makes a compromise between the volume of the hypersphere and violations of the boundary; and (2) One-Class Deep SVDD which is a simpler version that assumes that most of the training data is normal (i.e., low proportion of outliers).

The effectiveness of end-to-end approaches, and in particular end-to-end approaches optimized using a loss function with a direct anomaly detection objective, rather than one based on reconstruction error, motivates our proposed D-RBFDD network. Moreover, it is desireable for the anomaly detectors to be both interpretable and adaptable to new data and changing concepts of normality. However, these characteristics are not easily associated with SVMs \cite{ratsch2006learning,1318049,10.1145/3357384.3357816}, or approaches built upon an SVM foundation \cite{nguyen2018scalable}. On the other hand, because of their local learning approach, RBF networks easily lend themselves to interpretation \cite{sendhoff2000extracting, jin2003extracting}, and are adaptable to changing concepts \cite{liu2020fast,han2011efficient}. This makes a deep end-to-end anomaly detector based on RBF networks, such as D-RBFDD, an attractive approach. D-RBFDD is a fast, effective anomaly detector, trained in an end-to-end fashion and capable of learning latent representations directly aligned with an anomaly detection objective, and that lends itself to easy interpretation and adaptation to changing concepts of normality.  
\section{Deep Radial Basis Function Data Descriptor (D-RBFDD) Networks}\label{RBFDD networks}
This section describes the RBFDD network and three alternatives for adding depth to  these networks, the last of which we refer to as the Deep RBFDD (D-RBFDD) network. 
\subsection{Radial Basis Function Data Descriptor (RBFDD) Networks}

In our previous work \cite{RBFDD} we proposed the RBFDD network, which is a modification of the Radial Basis Function (RBF) network that enables it to be used for anomaly detection. An RBF network is a local-representation learning technique used for classification that divides the input space among a set of local kernels. In an RBF network, for every input data point, depending on where in the input space it appears, a fraction of these locally-tuned kernel units get activated. Activation is measured using a function of the distance between an input instance, \(X\), and the center, \(\mu_h\), of every kernel unit \(h\). Distance is typically measured with Euclidean distance, \norm{X - \mu_h}, and the activation function for the local kernels is usually implemented using a Gaussian function:
\begin{equation}
P_h(X) = \exp\left({-\frac{\norm{X - \mu_h}^2}{2s_h^2}}\right)
\end{equation}
\noindent where \(\mu_h\) and \(s_h\) denote the mean and standard deviation of the local unit \(h\). Activation is maximum when \(X = \mu_h\), and decreases as \(X\) and \(\mu_h\) diverge.

\begin{figure}[t]
        \begin{center}
            \includegraphics[scale=.65]{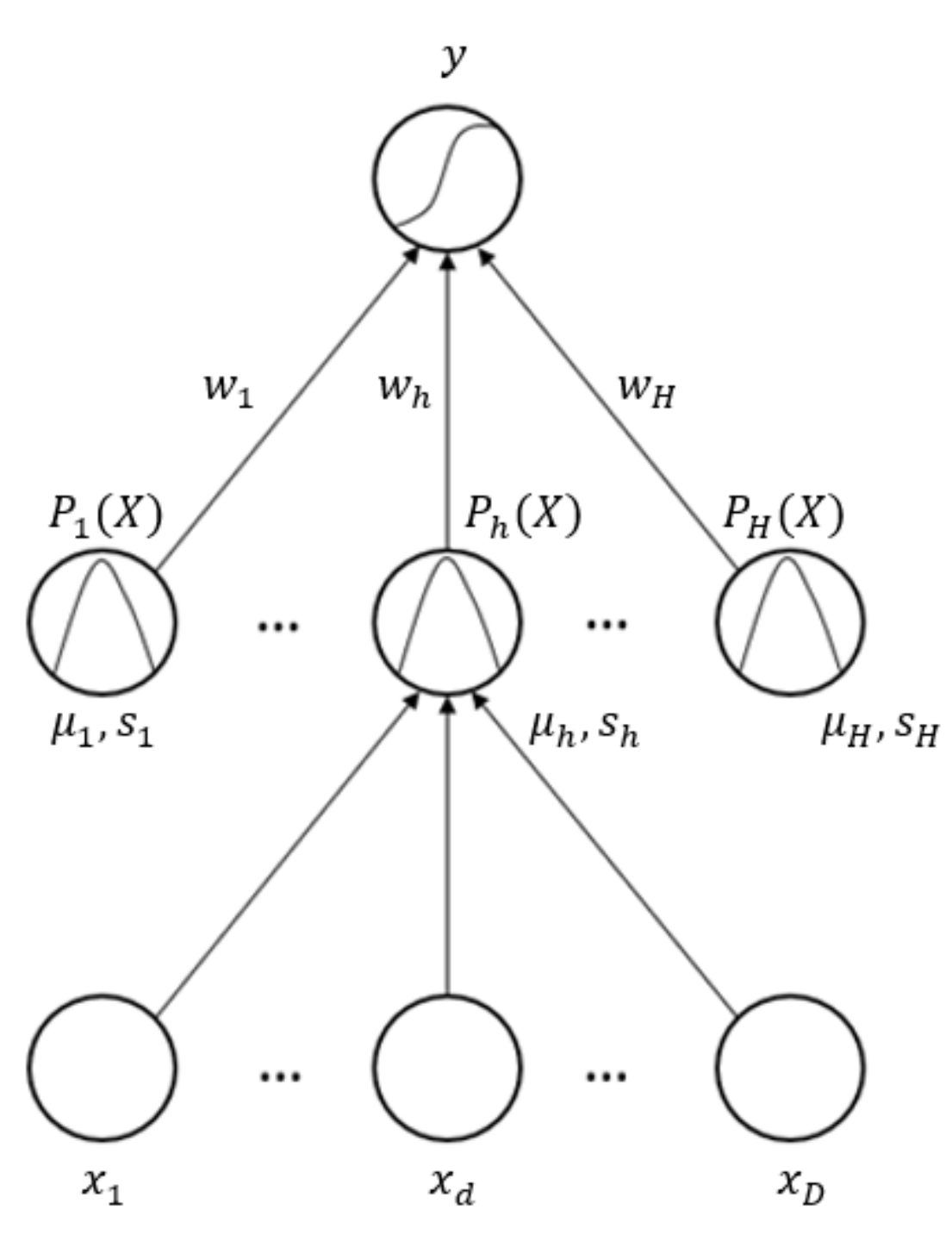}

            \caption{The RBFDD network.}
             \label{fig:RBFDD Network}
            
        \end{center}
\end{figure}

RBFDD networks adapt the RBF network approach to learn a set of Gaussian kernels that compactly represent normal instances in a training set, thus transforming them into anomaly detectors. A trained RBFDD network can be used as an anomaly detector by recognizing instances that are not covered by this compact representation. Figure~\ref{fig:RBFDD Network} shows the architecture of an RBFDD network. Here \(x_d\) denotes the \(d^{th}\) feature in the input data \(X\), which is a \(D\)-dimensional vector. In the output node of the network the \(tanh\) non-linear activation function proposed in \cite{LecunEfficientBackprop} is used, as it avoids saturation. In particular for a given \(D\)-dimensional input data, $X_i$, the output of the model is computed as: 
\begin{equation}\label{RBFDD output}
    y_i = 1.7159 \times tanh\left(\frac{2\times z(X_i)}{3}\right)  
\end{equation}
\begin{equation}\label{Z(X)}
    z(X_i) = \sum_{h=1}^{H} w_h \times P_h(X_i) 
\end{equation}
\noindent where $w_h$ is a weight connecting the Gaussian kernel $h$ to the output unit. 

After training, the output of the RBFDD network, \(y_i\), for a given input, \(X_i\), should be high if \(X_i\) belongs to the normal region of the input space and low otherwise. In the RBFDD network, the unsupervised pre-training phase used to train RBF networks \cite{Bishop} (i.e., \(k\)-means clustering \cite{Ethem}) remains in place. Following this step, the backpropagation of error algorithm is used with gradient descent to find the optimal values for the network parameters. In this process the following cost function is minimized for  mini-batches of size \(N\):
\begin{equation}
    \label{eq:RBFDD error}
     E = \sum\limits_{i=1}^{N}\left(\frac{1}{2}\left[\left(1 - y_i\right)^2 + \beta \sum\limits_{h=1}^{H} s_{h}^2 + \lambda \sum\limits_{h=1}^{H}{w_h^2} \right]\right)
\end{equation}

This cost function is a weighted summation of three terms. In the first term, \((1 - y_i)^2\), \(y_i\) is the output of the RBFDD network for a given \(D\)-dimensional input data instance, $X_i$. This term encourages the network to learn a model that outputs a value as close as possible to 1 for instances belonging to the normal class. The second term, regularizes the spreads of the Gaussian kernels in the hidden layer of the network, and is the squared L-2 norm \cite{GoodFellowDeepLearning} of the spreads for the \(H\) Gaussians. This encourages the most compact set of Gaussians possible to represent the normal data. The third term, regularizes the weights connecting the RBFDD hidden layer units to the output unit. This stops the weights from becoming so large that they would actually ignore the outputs from the hidden units, and makes the RBFDD network robust to outliers in the training set \cite{GoodFellowDeepLearning}. Minimizing Eq.\eqref{eq:RBFDD error}, using gradient descent, finds the most compact set of Gaussian kernels whose collective output is still high for the normal region of the input space and low everywhere else (i.e., where the anomalies are expected to appear). 
RBFDD networks use radial kernels, and thus, they might lack the necessary flexibility to learn certain distributions. To overcome this limitation of RBFDD, we previously proposed the Elliptical Basis Function Data Descriptor (EBFDD) \cite{bazarganielliptical}, where we make the anomaly detector more flexible by replacing the radial kernels with elliptical kernels. EBFDD was shown to perform better than RBFDD, however, at significantly increased computational cost (EBFDD requires a covariance matrix inversion which is a computationally very expensive operation). We believe, however, that the same flexibility can be achieved by adding computational layers before the RBFDD layer to transform the representation into a space where RBFDD can be applied effectively and, due to the lower computation time of RBFDD compared to EBFDD, also efficiently. Thus, we avoid the computational complexity of the EBFDD networks by adding more layers and retaining the capacity of the deep model to learn complex distributions in the normal data. Although the RBFDD network is an effective anomaly detector when used with well-behaved sets of input features, it does not perform well on high-dimensional raw data representations (e.g., pixel values in images or raw sensor data). This is the main motivation for deepening the structure of the RBFDD network so that we can apply it to anomaly detection problems with raw high-dimensional input data. The next section describes different alternatives for placing extra computational layers before the RBFDD network. 
\subsection{Deepening RBFDD Networks}\label{Deep RBFDD}

\begin{figure}[tbh]
\centering
\begin{subfigure}[b]{0.23\textwidth}
    \includegraphics[width=\textwidth]{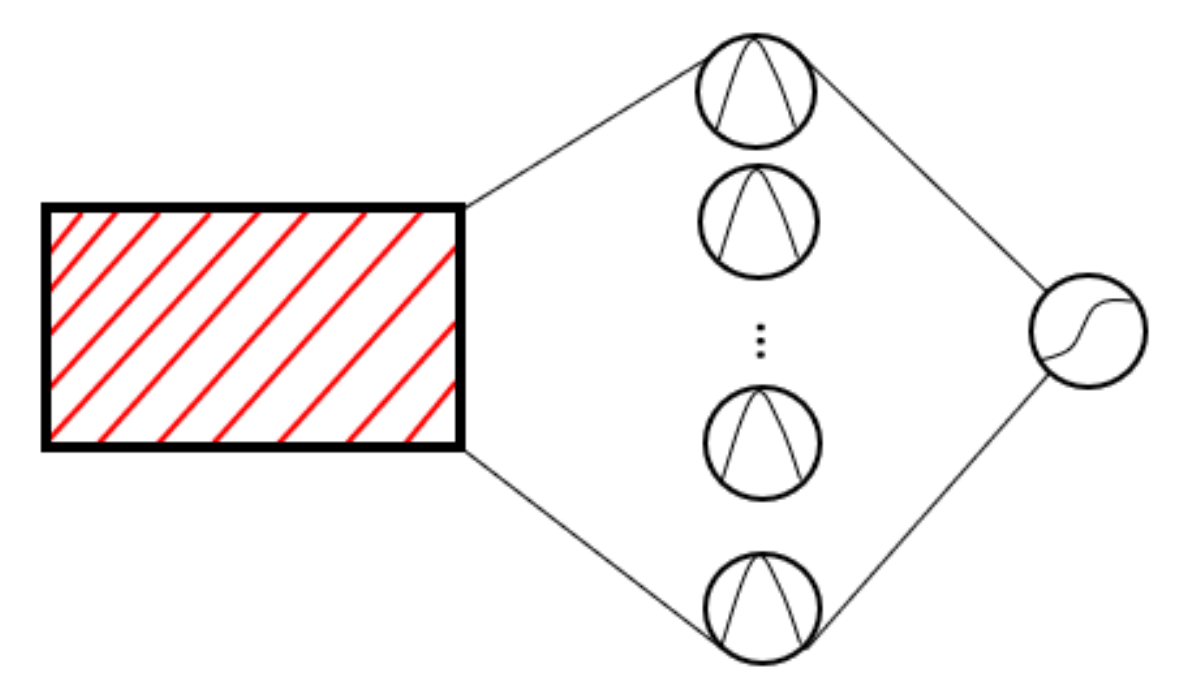}
    \caption{Fix-Res + RBFDD}
    \label{fig:Fixed ResNet-18 + RBFDD network}
\end{subfigure}
~ 
\begin{subfigure}[b]{0.23\textwidth}
    \includegraphics[width=\textwidth]{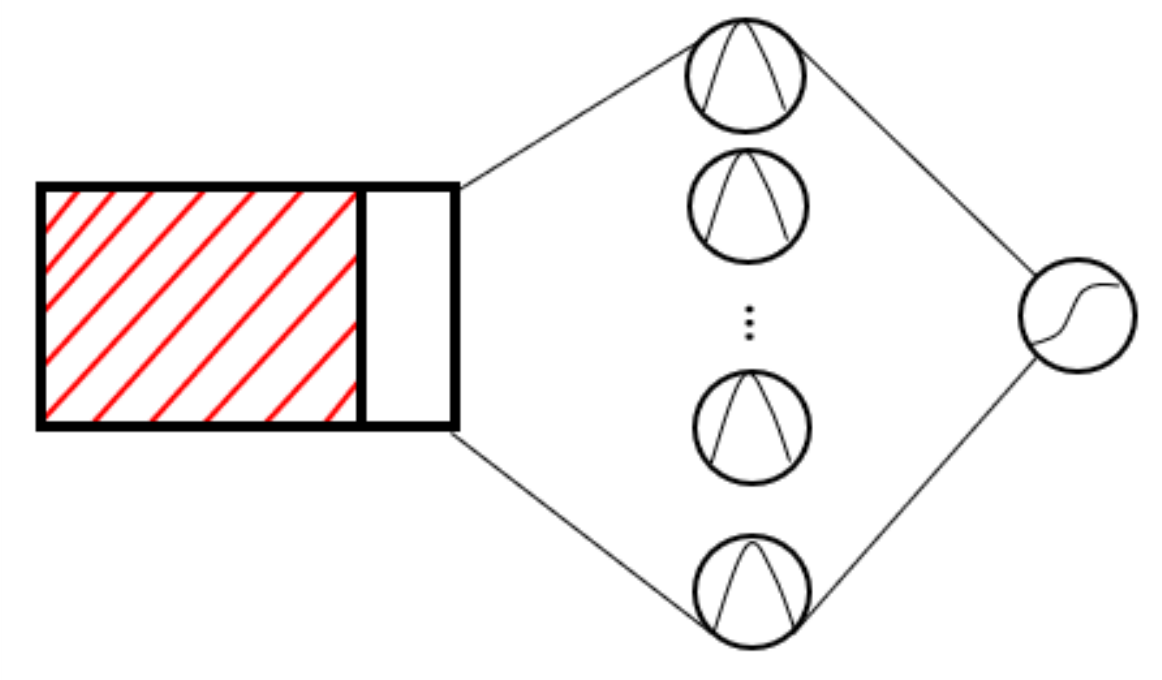}
    \caption{Fine-Res + RBFDD}
    \label{fig:Fine-tuned ResNet-18 + RBFDD network}
\end{subfigure}
~ 

\begin{subfigure}[b]{0.23\textwidth}
    \includegraphics[width=\textwidth]{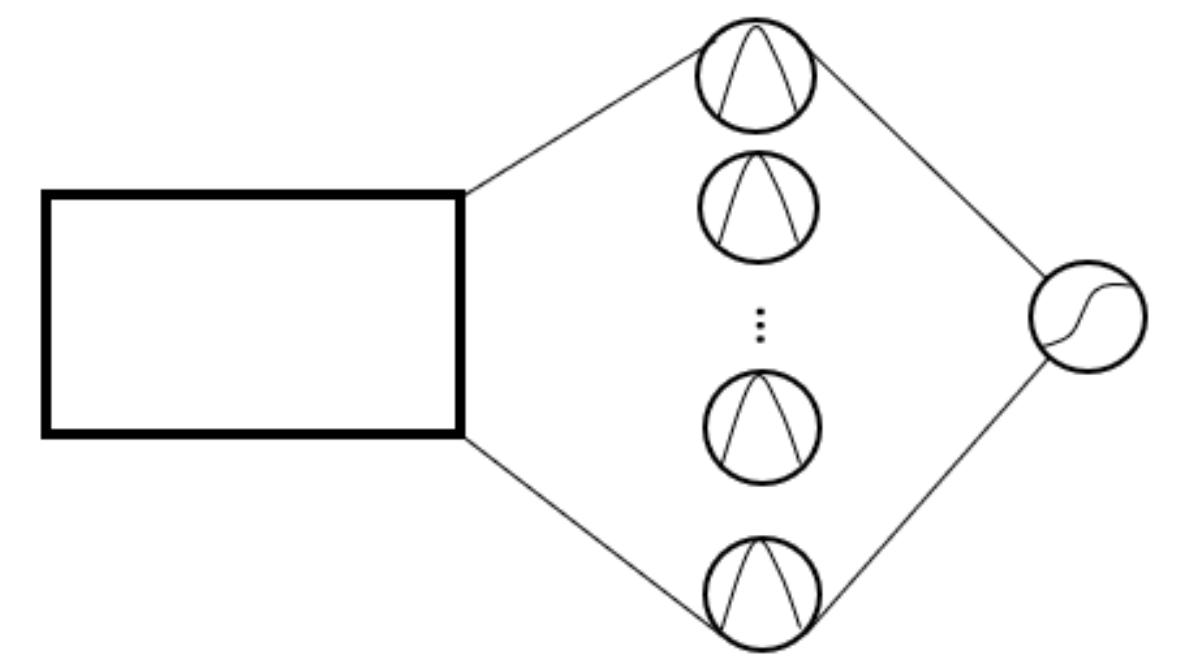}
    \caption{Deep RBFDD}
    \label{fig:LeNet-5 + RBFDD network}
\end{subfigure}
\caption{Three approaches to deepening RBFDD networks. The red lines highlight the fixed portion of each model.}\label{fig:Architectures of deep networks}
\end{figure}

We explore three ways to add depth to RBFDD networks (illustrated in Figure~\ref{fig:Architectures of deep networks}):
\begin{itemize}
\item \emph{Transfer learning} \cite{weiss2016survey} is exploited and the latent representation produced at the final layer of a network that is already trained on a large generic dataset is presented to the RBFDD network. 

\item The latent representation from a pre-trained network, such as that described above, can be \emph{fine-tuned} using the cost function optimized when training an RBFDD network.

\item Computational layers placed before the RBFDD network can be trained from random initialization as part of the overall \emph{end-to-end} deep RBFDD training process.
\end{itemize}

For the approach using transfer learning, referred to as \textit{Fix-Res + RBFDD}, we use a fixed, pre-trained ResNet-18 model trained on the ImageNet \cite{he2016deep} dataset, and extract the latent representation after its last convolutional layer for each data instance. This representation is then passed to a standard RBFDD model. In the second approach, referred to as \textit{Fine-Res + RBFDD}, again we connect a pre-trained ResNet-18 model to an RBFDD model. In this case, however, we fine-tune the last 4 convolutional layers, and the last 4 batch-norm layers of the pre-trained ResNet-18 model as part of training the RBFDD model using the cost function in Eq.\eqref{eq:RBFDD error}. This fine-tuning step is expected to make the latent representation passed to the RBFDD model more appropriate for anomaly detection, and improve its overall performance.  

\begin{figure*}[t]
        \label{D-RBFDD Architecture}
        \begin{center}
            \includegraphics[scale=.55]{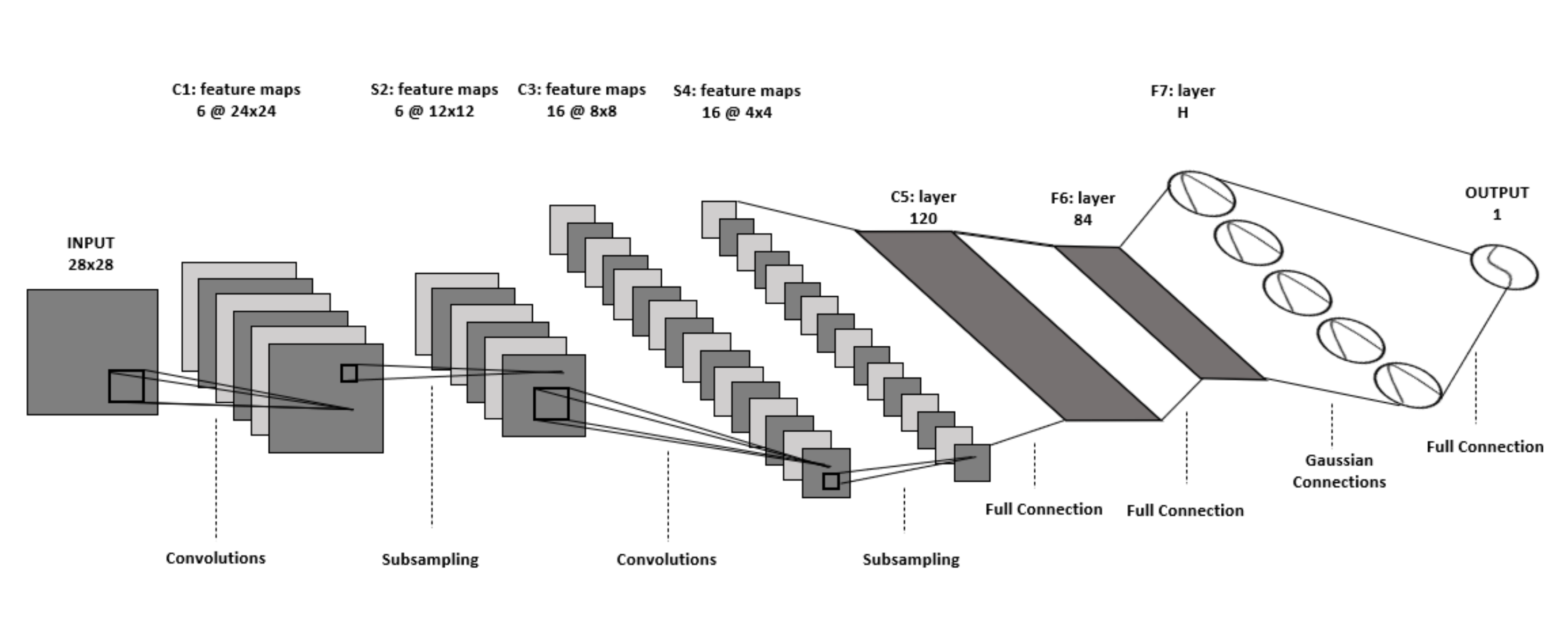}
            \caption{The Deep RBFDD Anomaly Detector}
            \label{fig:D-RBFDD}
        \end{center}
\end{figure*}

The final approach, that we refer to as the Deep RBFDD (D-RBFDD) network, attaches randomly initialized computational layers before the RBFDD layer and trains the entire network in an end-to-end fashion based on minimizing the cost function in Eq.\eqref{eq:RBFDD error}. Provided that the cost function can generate sufficient signal to train the entire deep model, the advantage of this method is that by using end-to-end training the latent representation that this network passes to the RBFDD layer will be more suited to anomaly detection than the representation generated by the pre-trained classification network, even when it is fine-tuned. In D-RBFDD we add layers following the simple LeNet-5 network architecture \cite{lecun1998gradient} preceding the RBFDD layer. The overall D-RBFDD network architecture is illustrated in Figure~\ref{fig:D-RBFDD}.\par

To facilitate the application of k-means clustering in the pre-training phase of training an RBFDD network we apply a \(tanh\) non-linear activation  \cite{LecunEfficientBackprop} to the latent representations coming into the RBFDD layer. This ensures that the k-means algorithm is provided with a bounded latent representation and leads to better model initialization.

Deepening the RBFDD network, should improve the performance of the RBFDD network and make it a stronger anomaly detector, in particular, when dealing with raw high-dimensional input data.
\section{Experimental Setup}\label{Experiment Setup}
We have designed an experiment to evaluate the performance of the different approaches to deepening RBFDD networks, and to compare these to state-of-the-art anomaly detection approaches\footnote{The code for the D-RBFDD network is available on GitHub repository: https://github.com/MLDawn/DRBFDD}.  
We include the following state-of-the-art anomaly detection approaches in this experiment: One-class Deep Support Vector Data Descriptor (DeepSVDD-OC) networks, Soft-boundary Deep Support Vector Data Descriptor (DeepSVDD-SB) networks, One-Class Support Vector Machines (OCSVMs), Isolation Forest (iForest), RBFDD networks, and deep Convolutional Auto Encoders (CAEs). In all cases only \emph{normal} data is used during model training.

To further investigate how effectively representations can be transferred from pre-trained classification networks to anomaly detection tasks, in the case of the OCSVM and iForest models (as well as RBFDD networks), we have also considered the scenario where the latent representation learned by a pre-trained classification network is used as input. We also use the latent representation learned by the version of RBFDD that fine-tunes the pre-trained classification model representation as input to these shallow models to better understand the impact of fine-tuning.

\subsection{Datasets \& Anomaly Detection Scenarios}\label{datasets}
We use two well-known labelled image classification datasets, MNIST \cite{lecun-mnisthandwrittendigit-2010} and Fashion MNIST \cite{xiao2017/online}, to explore the effectiveness of the three approaches to deepening RBFDD networks.
We generate multiple anomaly detection scenarios using these datasets. In each scenario we consider one class as normal and another class as anomalous. These pairs of classes (shown in Table \ref{Experiment Results on the classification datasets}) were selected to cover simple and challenging scenarios. For example, for MNIST we have a simple scenario where digit 0 is considered normal and digit 1 is anomalous, and similarly for Fashion MNIST we have  a scenario where \emph{T-shirts/tops} are normal and \emph{boots} are anomalous. Images from these pairs of classes are easily discernible, and we expect to see high performance across most of the algorithms. We also have more challenging scenarios. For instance, from MNIST we have a scenario where digit 4 is normal and digit 9 is anomalous, and for Fashion MNIST we have a scenario where \emph{T-shirts/tops} are normal and \emph{shirts} are anomalous. These pairs of images are not easy to separate as they are so similar.

We also use the MIT-BIH Arrhythmia Database\footnote{https://physionet.org/content/mitdb/1.0.0/} \cite{932724,goldberger2000physiobank}, a real-world  highly-imbalanced anomaly detection timeseries dataset, to compare the performance of the D-RBFDD network with  state-of-the-art algorithms. This dataset contains 47 half-hour excerpts of two-channel ambulatory ECG recordings obtained from 48 subjects. We pre-processed the data  to reduce the dimensionality by down-sampling from 360Hz to 187Hz and to extract individual heart-beats, each of which has an associated ground-truth label in the dataset\footnote{The peaks of each heartbeat are labelled in this dataset. Following the approach in \cite{xu2018towards}, we considered the mid-point between every two consecutive peak values to be the border between two consecutive heart-beats. All extracted heart-beats are zero-padded/truncated to a length that is higher than 95\% of the extracted heart-beat lengths, that is, 417.}. Out of the 19 anomalous classes in the dataset, we use the 4 most common to make four binary anomaly detection scenarios, and a \emph{One vs. All} scenario in which all of these 4 anomalous classes are considered together.


\begin{table*}[!htb]
\centering
    \caption{Experiment results on the MNIST and Fashion MNIST datasets. Each column is labelled N-A, where N = normal class and A = anomalous class. The values in each cell are AUC scores followed by relative rank in parentheses. The average rank per algorithm is given in the last column. The labels for Fashion MNIST are: T: T-shirts/tops, B: Ankle boots, S:Shirts, Sn:Sneakers and Sa:Sandals.}
 \label{Experiment Results on the classification datasets}
\adjustbox{max width=\textwidth}{
     {\renewcommand{\arraystretch}{1.2}%
    \begin{tabular}{l |r r r r r r| r r r r | r}
        \hline 
          & \multicolumn{6}{|c|}{MNIST} & \multicolumn{4}{c|}{Fashion MNIST}& 
         \tabularnewline 
          & 0 - 1 & 7 - 1 & 4 - 9 & 7 - 9 &9 - 4& 9 - 7& T- B & T - S & Sa - Sn &  B - Sa & Avg. Rank\tabularnewline
         \hline 

        iForest  & 0.9648 (\,~7)  &  0.7725 (12)& 0.6296 (\,~7)& 0.7948 (\,~5)&  0.7484 (\,~6)& 0.7355 (\,~6)& 0.9963 (\,~3)& \textbf{0.8182} (1)& 0.5394 (10)& 0.9536 (\,~7) & 6.4 (\,~6)\tabularnewline
        
        Fix-Res + iForest  & 0.4795 (14)& 0.4608 (14)& 0.5904 (12)& 0.5939 (14)&0.5595 (14)& 0.5941 (14)& 0.6212 (14)&0.5614 (13)& 0.5304 (11)& 0.4782 (14) & 13.4 (14)\tabularnewline

        Fine-Res + iForest  & 0.7609 (11)& 0.9698 (\,~5)& 0.5807 (14)& 0.6910 (\,~9)& 0.6898 (10)& 0.6566 (11)& 0.9761 (10)& 0.4585 (14)& 0.4957 (14)&  0.7753 (10) & 10.8 (12)\tabularnewline
        \hline
        
        OCSVM  & 0.9962 (\,~3)& 0.9623 (\,~6)& 0.8320 (\,~2)& \textbf{0.9209} (\,~1)& 0.9245 (\,~2)& 0.9125 (\,~2)& 0.9967 (\,~2)& 0.7872 (\,~6)& 0.5708 (\,~9)&  0.9708 (\,~5) & 3.8 (\,~3)\tabularnewline

        Fix-Res + OCSVM & 0.5506 (13)& 0.6924 (13)& 0.5808 (13)&0.6098 (13)& 0.6035 (11)&  0.6268 (13)& 0.8981 (12)& 0.5992 (12)& 0.5100 (12)& 0.6011 (13) & 12.5 (13)\tabularnewline

        Fine-Res + OCSVM & 0.8542 (\,~9)& 0.9746 (\,~3)& 0.5975 (11)& 0.7181 (\,~7) &0.7119 (\,~8)& 0.6850 (\,~9)& 0.9780 (\,~8)& 0.6429 (11)& 0.5036 (13)& 0.7232 (12) & 9.1 (\,~9)\tabularnewline
        
        \hline

        RBFDD  & \textbf{0.9988} (\,~1)& 0.9722 (\,~4)&  0.7585 (\,~3)& 0.8187 (\,~4)& 0.8069 (\,~5)& 0.8583 (\,~5)& 0.9954 (\,~4)& 0.7898 (\,~5)&  0.6562 (\,~6)& 0.9830 (\,~4) & 4.1 (\,~4)\tabularnewline

        Fix-Res + RBFDD  & 0.7923 (10)& 0.8332 (11)& 0.6273 (8)& 0.6157 (12)& 0.6941 (\,~9)& 0.6714 (10)& 0.9740 (11)& 0.6881 (9)&  0.6197 (8)& 0.7299 (11) & 9.9 (11)\tabularnewline

        Fine-Res + RBFDD& 0.9422 (\,~8)& 0.9119 (\,~9)& 0.6217 (\,~9)& 0.6612 (11)&  0.7236 (\,~7)& 0.7152 (\,~7)&  0.9901 (\,~7)& 0.8055 (\,~2)& \textbf{0.7310} (\,~1)& 0.8431 (\,~9) & 7.0 (\,~7)\tabularnewline
        \hline
        
        CAE-1  & 0.7000 (12)& 0.8964 (10)& 0.7137 (\,~6)& 0.8934 (\,~3)& 0.5688 (13)& 0.6419 (12)& 0.8426 (13)& 0.6823 (10)&0.6379 (\,~7)&0.8875 (\,~8) & 9.4 (10)
         \tabularnewline

        CAE-2  & 0.9914 (\,~5)& 0.9514 (\,~7)&  0.6110 (10)& 0.6614 (10)& 0.5843 (12)& 0.7139 (\,~8)& 0.9764 (\,~9)&0.7378 (\,~8)&0.7013 (\,~4)& 0.9647 (\,~6) & 7.9 (\,~8)\tabularnewline
        \hline

         DeepSVDD-OC  & 0.9906 (\,~6)& \textbf{0.9943} (\,~1)& 0.7455 (\,~4)& 0.7071 (\,~8)& 0.9140 (\,~3)& 0.8950 (\,~4)& 0.9950 (\,~6)& 0.8001 (\,~4)& 0.6567 (\,~5) & 0.9871 (\,~3) & 4.4 (\,~5)\tabularnewline

        DeepSVDD-SB & 0.9957 (\,~4)& 0.9916 (\,~2)& 0.7365 (\,~5)& 0.7417 (\,~6)& 0.9132 (\,~4)& 0.8969 (\,~3)& 0.9951 (\,~5)& 0.8020 (\,~3)& 0.7023 (\,~3)& \textbf{0.9896} (\,~1) & 3.6 (\,~2) \tabularnewline

        \hline

        D-RBFDD & 0.9981 (\,~2)& 0.9512 (\,~8)& \textbf{0.8450} (\,~1)& 0.8971 (\,~2)& \textbf{0.9480} (\,~1)& \textbf{0.9137} (\,~1)& \textbf{0.9987} (\,~1)& 0.7459 (\,~7)& 0.7161 (\,~2)& 0.9887 (\,~2) & \textbf{2.7} (\,~1)\tabularnewline
        
        \hline
        
    \end{tabular}}
    }

\end{table*}

\begin{table*}[ht]
\centering \caption{Experiment results on the MIT-BIH Arrhythmia dataset. The label of each anomalous class is given at the top of the columns (for the label descriptions see supplementary materials). The values of each cell are AUC scores followed by the relative rank in parentheses. The average rank per algorithm is given in the last column.}\label{Experiment Results on Anomaly Detection Datasets}
 %
{\renewcommand{\arraystretch}{1.0}%
\begin{tabular}{l r r r r r| r}
\hline 
 & L  & R  & V  & /  & One vs. All &Avg. Rank\tabularnewline
        \hline 
        iForest & 0.5743 (8)& 0.7118 (8) & 0.6819 (8) & 0.7713 (8)&  0.6808 (8) & 8.0 (8)\tabularnewline
        \hline 
        OCSVM & 0.6684 (5) & 0.7582 (6) & 0.8647 (5)& 0.8591 (6)& 0.7830 (5) & 5.4 (5)\tabularnewline
        \hline 
        RBFDD & 0.7002 (4) & 0.8182 (4)& 0.8722 (4)& 0.8947 (5)& 0.8043 (4) & 4.2 (4)\tabularnewline
         \hline 
        CAE-1 & 0.6331 (6) & 0.7525 (7)& 0.7416 (6)& 0.8139 (7)& 0.7174 (6) & 6.4 (7)\tabularnewline
        CAE-2 & 0.5994 (7) & 0.8139 (5)& 0.7407 (7)& 0.9395 (1)& 0.7023 (7) & 5.4 (5)\tabularnewline
        \hline
        DeepSVDD-OC & 0.7700 (3)& 0.8352 (3)& 0.9241 (3)& 0.9187 (4) &  0.8324 (3) & 3.2 (3)\tabularnewline
        
        DeepSVDD-SB &  0.7835 (1) & 0.8596 (1)& 0.9363 (1)& 0.9316 (2) & 0.8346 (2) & 1.4 (1)
        \tabularnewline
        \hline
        D-RBFDD & 0.7723 (2) & 0.8458 (2)& 0.9361 (2)& 0.9261 (3)& 0.8507 (1) & 2.0 (2)\tabularnewline
        \hline
    \end{tabular}}
\end{table*}


In all experiments, the models are trained using \textit{only} instances of the normal class, and during testing we provide unseen samples from both normal and anomalous classes to measure the performance of the different models. For all the datasets, feature values have been normalized to $[0,1]$. In particular, the normalization for both the MNIST and Fashion MNIST datasets is done by dividing individual pixel values by 255, as this is the maximum pixel value for grey-scale images. For the MIT-BIH Arrhythmia dataset, the sample values range from 0 to 2047 inclusive, with 1024 (i.e., the mid-point) corresponding to zero. This is due to the fact that, at the digitization step, a resolution of 11-bits has been used, resulting in $2^{11}$ levels, which are the actual resultant values of the signal in this dataset. Thus, normalization is done by dividing individual values in the ECG signals by 2047\footnote{Since, the original value of 1024 will be translated to 0.50 in the normalized space, then 0.50 is the value by which we will pad at the end of our signals after segmenting the heart-beats. In terms of truncating, we will simply truncate the ending of the signals, where the heart-beat has a length more than 417, and this is for the heart-beats that have a length in the top 5\% of the heart-beat lengths.}.

The different datasets and the scenarios used are summarized in the supplementary material.

\subsection{Experimental Design}\label{sec:experiment_design}
To evaluate models we use an approach based on  bootstrapping that makes maximum use of the anomalous samples available. For each iteration we randomly select 80\% of all normal instances in the dataset (with no replacement) to train the model. The remaining 20\% of normal instances is then mixed with all of the anomalous instances to form the test set. We perform 10 iterations of training and testing in this way and measure the area under the ROC curve (AUC) on the test set for each one. The AUC scores are then averaged across iterations to give the overall model performance. 

We perform hyper-parameter tuning using a grid search that repeats the above process for each hyper-parameter combination. The range of searched hyper-parameters are listed in supplementary material.
We report the best averaged AUC from the grid search for the corresponding experiment. We are aware that reporting the performance of the  models with the best set of hyper-parameters over-estimates the generalization performance of the models (known as the problem of \emph{many comparisons in induction algorithms} \cite{10.1007/978-3-319-07064-3_1}). However, as our goal is a relative comparison of algorithms, rather than an absolute estimate of generalization error, and all algorithms are evaluated in the same way this is an appropriate evaluation approach that makes better use of limited anomalous samples than measuring performance on a separate hold-out test set.

\subsection{State-of-the-Art Approaches}\label{state of the art}
Each state-of-the-art approach compared in this experiment is tuned to achieve its best possible performance (full details are provided in the supplementary material). For all OCSVM models we use Gaussian kernels, as recommended in \cite{DBLP:journals/corr/KhanM13}. The hyper-parameters tuned for OCSVM models are \(\nu\), and \(\gamma\). For iForest, the only hyper-parameter to be tuned is the number of estimators.

To explore their performance we use different CAE architectures, each with similar capacity to the D-RBFDD model. For the image classification datasets CAE-1 has two 2D  convolutional layers in the encoder and two transposed 2D convolutional layers in the decoder, while CAE-2  has three convolutional layers in the encoder and three transposed convolutional layers in the decoder. For the ECG dataset,  CAE-1 has two 1D convolutional layers in the encoder and two transposed 1D convolutional layers in the decoder.  CAE-2 has the same structure but the second 1D convolutional layer has twice the number of convolutional filters compared to CAE-1. For all CAEs rectified linear activation functions and max-pooling are used at each layer.

For the RBFDD network and the D-RBFDD network, the hyper-parameters that are tuned are the number of Gaussians in the hidden layer, and the coefficients of the cost function (Eq. \eqref{eq:RBFDD error}): \(\beta\) and \(\lambda\) (whose values fall in the range of (0, 1]). The D-RBFDD network, is based on the LeNet-5 network architecture \cite{lecun1998gradient}. For the ECG dataset we replace the 2D convolutions with 1D convolutions.

In the case of DeepSVDD, a LeNet-type network architecture is used \cite{pmlr-v80-ruff18a}, which we use for the image datasets. For the ECG dataset we replace this with the 1D LeNet-5 architecture used in the D-RBFDD network. In both versions of DeepSVDD the weight decay coefficient $\lambda$ is a tuned hyper-parameter. For DeepSVDD-SB, $\nu$, is also a tunable hyper-parameter, whose role is to control the trade-off between violations of the boundary and the volume of the hypersphere. Following the training method in \cite{pmlr-v80-ruff18a}, the training of DeepSVDD models also includes a learning rate scheduler that reduces the learning rate by a factor of 10 after a 75\% of the specified training epochs have been completed. 

In the experiments using the image classification datasets we use a pre-trained ResNet-18 model \cite{he2016deep} trained on the ImageNet \cite{deng2009imagenet} dataset\footnote{The ResNet-18 implementation used is available at: https://github.com/pytorch/vision/tree/master/torchvision/\\models/resnet.py} for transfer learning. No transfer learning is used for the ECG dataset, as reliable large-scale pre-trained models for ECG sensor data are not available.
\section{Results and Discussions}\label{Results and Discussions}
The results of the experiments based on the image classification datasets are detailed in  Table~\ref{Experiment Results on the classification datasets}. These results were achieved using the best hyper-parameter combinations found during the grid search described in Section \ref{sec:experiment_design} (these are listed in the supplementary material). 
For each anomaly detection scenario the different approaches have been ranked and the average ranks for each approach are summarized in the last column of each table. These results show the effectiveness of deepening RBFDD for raw datasets, and allow us to compare the three different strategies for deepening described in Section \ref{RBFDD networks}. The fact that the D-RBFDD model has out-performed the RBFDD model, in the majority of cases, demonstrates the value of using the deep model to generate a latent representation suitable for use by RBFDD. Moreover, it is interesting to note that none of the models that use the latent representation output by the fixed, pre-trained ResNet-18 model out-perform their equivalent models trained on the raw, high-dimensional representation. This is the case for the RBFDD models as well as for the OCSVM and iForest models. This is a reminder of the issue with \textit{mixed approaches} for deep anomaly detection mentioned in Section \ref{Related Work}. The fixed pre-trained ResNet-18 model has been trained for a multi-class classification objective and the latent representations generated by this network seem to be too entangled with that task to be very useful for anomaly detection. 

This is further underlined by the fact that, in almost all cases, the performance of the models (RBFDD, OCSVM, and iForest) using the latent representations that arise from the version of the ResNet-18 model fine-tuned using the RBFDD network improves over the versions of the models trained using the representations from the fixed ResNet-18 model. However, it is important to note that in most cases this performance was still not better than the models working on raw data.

Together these results show that deepening RBFDD networks allows them to work effectively with raw inputs, and that the D-RBFDD approach is the best way to do this out of those compared. This conclusion is reinforced by the results based on the ECG dataset shown in Table~\ref{Experiment Results on Anomaly Detection Datasets}. In experiments using this dataset D-RBFDD outperforms RBFDD in all cases.

By examining the results for the image classification datasets in Table~\ref{Experiment Results on the classification datasets} and those based on the ECG dataset in Table~\ref{Experiment Results on Anomaly Detection Datasets} together we can evaluate how D-RBFDD compares to other state-of-the-art anomaly detection algorithms. 
In the image classification dataset cases, the results show that, overall, the D-RBFDD network out-performs the other algorithms as it has the lowest average rank (lower ranks are better). On the ECG dataset DeepSVDD-SB has a slightly better average rank than D-RBFDD, although D-RBFDD performs better in the \emph{One vs. All} scenario which is  particularly important for anomaly detection as it is likely that anomalies will arise from very different data distributions.

\begin{figure}[t]
    \begin{subfigure}[t]{0.45\textwidth}
    \centering
    \includegraphics[width=\textwidth]{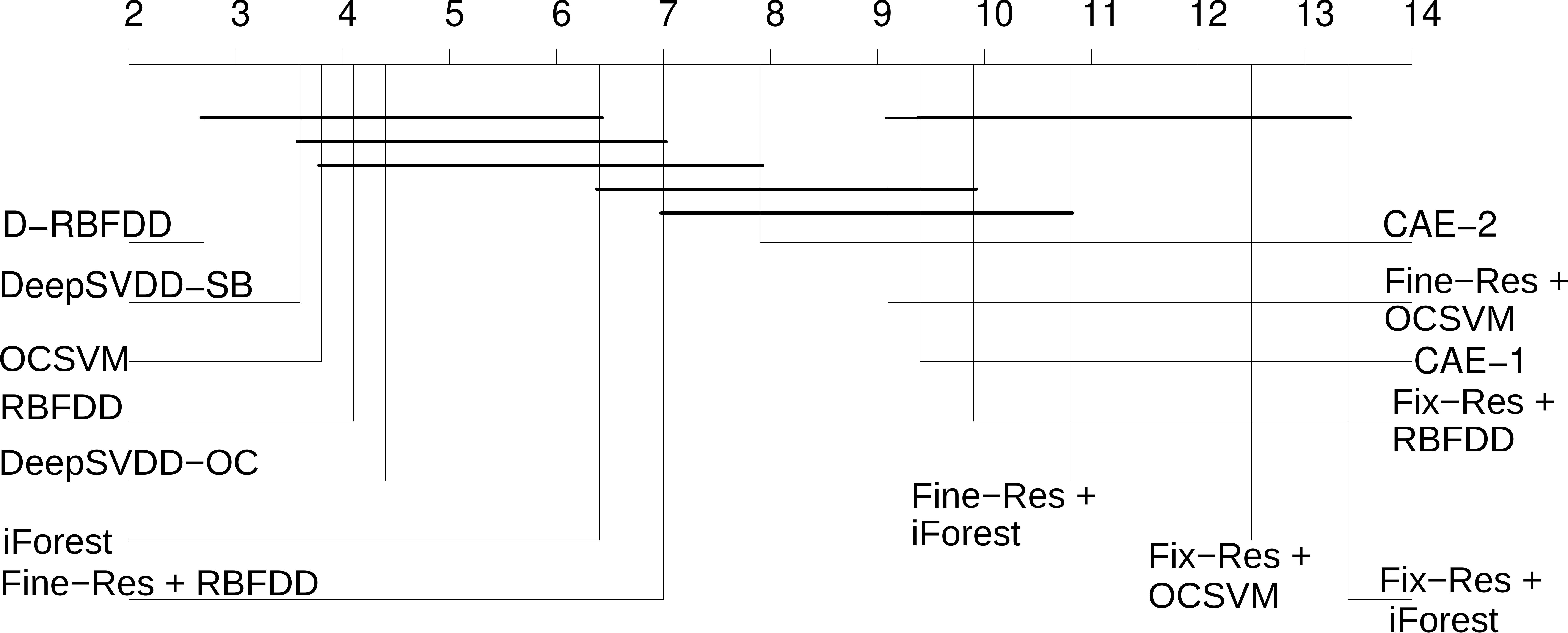}
    \caption{On cases of MNIST and Fashion MNIST datasets}
    \label{fig:cdplot_mnist}
    \end{subfigure}\hfill\begin{subfigure}[t]{0.45\textwidth}\includegraphics[width=\textwidth]{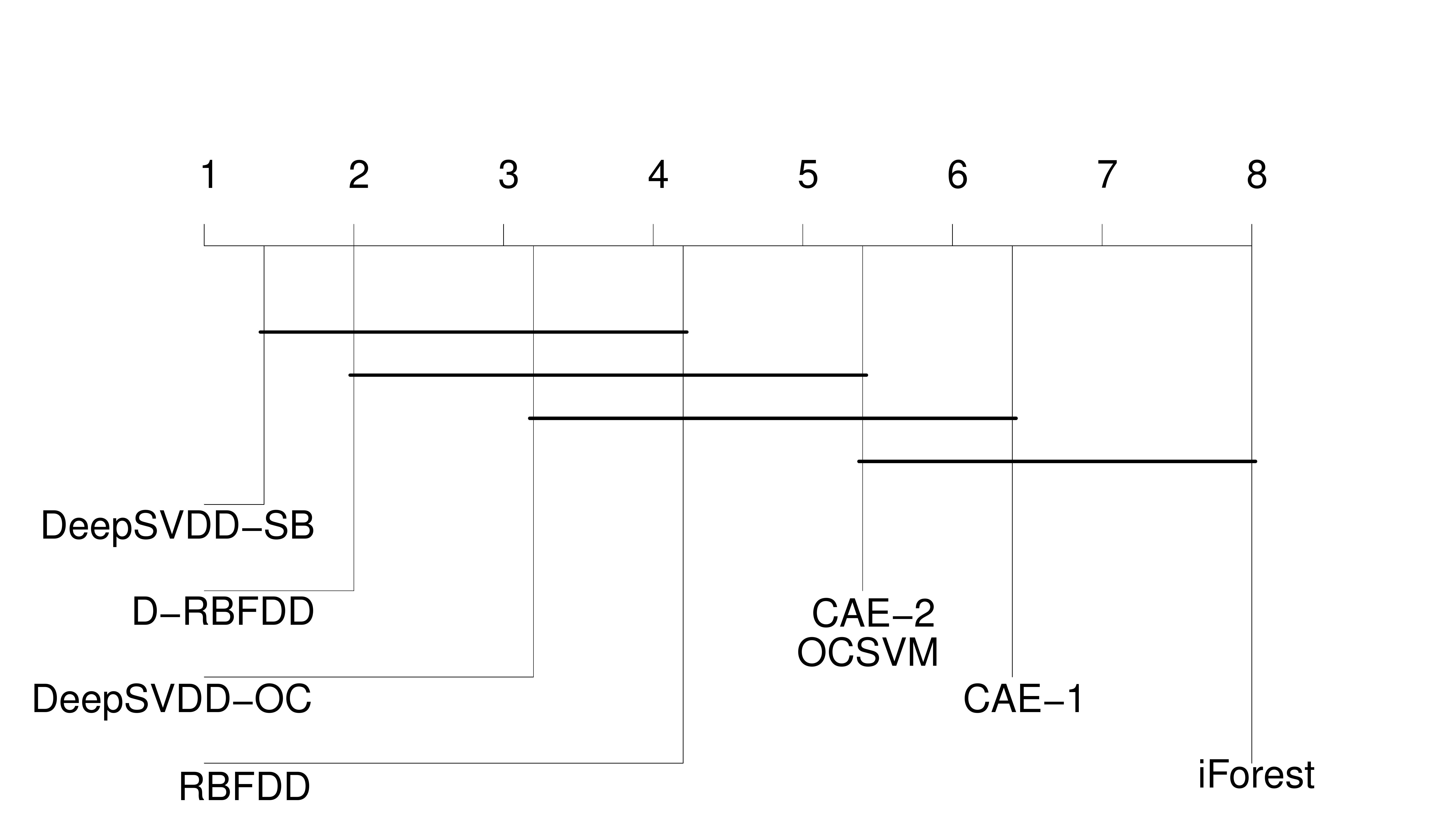}
    \caption{On cases of MIT-BIH Arrhythmia dataset}
    \label{fig:cdplot_ecg}
    \end{subfigure}
    \caption{Critical difference plots from a Friedman test using a significance level of $0.05$ on the different anomaly detection scenarios. Algorithms \emph{not} connected with horizontal bars are significantly different.}
\end{figure}

To further investigate and understand the overall differences between the performances of the different algorithms and the effectiveness of D-RBFDD, we perform non-parametric statistical significance tests for multiple classifier comparison. Following \cite{GARCIA20102044} a Friedman test followed by a Finner $p$-value correction was performed on the results in both Table~\ref{Experiment Results on the classification datasets} and \ref{Experiment Results on Anomaly Detection Datasets}. This test will analyze the difference in performance between each pair of algorithms with respect to the different anomaly detection scenarios.

The statistical test results for the image classification and ECG datasets are summarized in the critical difference plots (with significance level of $\alpha = 0.05$) in Figure~\ref{fig:cdplot_mnist} and Figure~\ref{fig:cdplot_ecg} respectively. Two algorithms not connected with bold horizontal lines are significantly different. The $p$-values of the statistical tests and the pairwise win/lose/tie results are provided in the supplementary material.




Figure~\ref{fig:cdplot_mnist} shows that D-RBFDD performs significantly and consistently better than the following algorithms: Fine-Res + RBFDD, CAE-2, Fine-Res + OCSVM, CAE-1, Fix-Res + RBFDD, Fix-Res + OCSVM, Fix-Res + iForest, Fine-Res + iForest. In the case of DeepSVDD-SB, OCSVM, RBFDD, DeepSVDD-OC and iForest, the null-hypothesis of the test could not be rejected with a significance level of $\alpha=0.05$, but D-RBFDD performed better in average rank. On the other hand, in a simple and direct pairwise win/lose/tie based comparison, D-RBFDD won in at least 70\% and up to 100\% of the anomaly detection cases when compared to the other algorithms (see supplementary material). This indicates that D-RBFDD typically performs as good as or better than the benchmark algorithms it is compared to. 

From Figure~\ref{fig:cdplot_ecg} we can see that DeepSVDD-SB attained the best average rank of 1.4 on the ECG dataset. D-RBFDD achieved similar performance with an average rank of 2.0. In all of the scenarios DeepSVDD-SB has performed slightly better than D-RBFDD, but, interestingly D-RBFDD achieved the best performance in the \emph{One vs. All} case. Although, from an overall point of view DeepSVDD-SB and D-RBFDD, the null hypothesis could not be rejected in any of the significance levels. D-RBFDD performed better than iForest and CAE-1 at the significance level of $\alpha=0.01$ and $\alpha=0.05$ respectively, and performed better than OCSVM and CAE-2 with a significance level of $\alpha=0.1$.




Overall these results indicate that adding extra computational layers to RBFDD makes it a much more effective anomaly detector for problems with raw data representations.
Also, selecting D-RBFDD will lead to at least similar or many times better performance than the other approaches, making it an attractive solution for anomaly detection for these types of datasets. We believe that this strong performance, coupled with the easy interpretability and adaptability of approaches based on RBF networks make D-RBFDD a compelling approach. 
\section{Conclusions \& Future Work}\label{Conclusion and Futurework}
In this article, we have proposed a deep one-class neural network, the D-RBFDD network, that adds convolutional layers before an RBFDD network. The D-RBFDD network is trained in an end-to-end fashion on an objective that is designed specifically for anomaly detection. We have shown that this network has successfully turned the shallow RBFDD network into a deep one-class classifier, suitable for anomaly detection on high-dimensional raw data such as images and sensor data. 
Unlike some of the state-of-the-art algorithms, in particular OCSVMs, the D-RBFDD networks are scalable in size and can work with large datasets and high-dimensional data.

In a set of benchmark experiments, for image datasets, the D-RBFDD network has shown superior performance to state-of-the-art one-class classifiers---DeepSVDD, OCSVM, iForest, and CAEs. In the case of the ECG dataset, the D-RBFDD network has produced competitive results to those of  the DeepSVDD-SB algorithm, and out-performed it in the \emph{One vs. All} scenario, which is  particularly important for anomaly detection as it is likely that anomalies will arise from very different data distributions. We have also observed that our proposed D-RBFDD model, has indeed out-performed its shallower version, the RBFDD network, in almost all of our benchmark experiments. This suggests that, when dealing with raw data we have benefited from the introduction of depth in the D-RBFDD network.

Our experiments show that transfer learning using a pre-trained ResNet-18 with fixed weights, does not work well for anomaly detection. We believe that the reason for this is due to the fact that ResNet-18 has been trained for a multi-class classification task and that the latent representations that it generates are too entangled with that task to be useful for anomaly detection. Interestingly, we see that if the final layers of the ResNet-18 model are fine-tuned using the RBFDD network cost function, the performance improves in most cases. 

Overall, it can be concluded that, selecting D-RBFDD for an anomaly detection task on raw data would lead to performance that is at least as good as or significantly better than current state-of-the-art algorithms.

The D-RBFDD network is an attractive option for the task of anomaly detection. First of all, it is showing significant improvement over its predecessor the RBFDD network and it is showing competitive performance with  state-of-the-art anomaly detection algorithms. This is a pre-requisite for broadening the application of such networks to more challenging scenarios such as learning from streams of incoming data, where the main challenge is the dynamic nature of what constitutes normal and anomalous, known as, concept drift. In a D-RBFDD network, we have the ability to control the number of Gaussians, which would equip the network with a high degree of adaptability for scenarios where concept drift is a concern and the definition of normal would change over time. Thus, by adding/removing/replacing Gaussians, the D-RBFDD network could learn a variety of new emerging contexts as well as forget the expired ones. Furthermore, D-RBFDD networks have the potential to give us a reasonable degree of interpretability as to why an input is flagged as an anomaly. The features learned by the RBFDD component in the D-RBFDD network (i.e., centers and spreads of Gaussian kernels and associated weights) provide us with a level of interpretability that has potential to be quite informative in terms of understanding the model learned and the reasoning behind flagging anomalies.

In the future we plan to exploit the flexibility of D-RBFDD to adapt it for an on-line learning scenario where detection and handling concept drift in the incoming stream of data is important. We will explore approaches to allow the D-RBFDD network, to self-expand and prune to adapt to the appearance or disappearance of concepts.

\bibliographystyle{IEEEtran}
\bibliography{main}

\begin{IEEEbiographynophoto}{Mehran H. Z. Bazargani}
 received his B.Sc. degree in Information Technology from University College of Nabi Akram (UCNA), Iran, 2010, and his M.Sc. degree in computer engineering from Easterm Mediterranean University (EMU), Cyprus, in 2013. He is currently in his last semester of Ph.D. in computer science with the Insight Centre for Data Analytics, at University College Dublin (UCD), Ireland. He is the founder of the machine learning educational platform: MLDawn (www.mldawn.com). His research interests include machine learning, deep learning, anomaly detection, and timeseries analysis.
\end{IEEEbiographynophoto}

\begin{IEEEbiographynophoto}{Arjun Pakrashi} received the B.Sc. degree (Hons.) in computer science from Calcutta University (CU), India, in 2011, the master's degree in computer science from Banaras Hindu University (BHU), India, in 2013, and the Ph.D. degree in computer science from University College Dublin, in 2020. He worked in the industry until 2015. He is currently a Postdoctoral Researcher with the Insight Centre for Data Analytics, University College Dublin. His research interests include machine learning, multi-label classification, ensemble methods, and anomaly detection.
\end{IEEEbiographynophoto}

\begin{IEEEbiographynophoto}{Brian Mac Namee} received the Ph.D. degree in computer science from Trinity College Dublin, Ireland, in 2004. After a period working in the industry and at the Dublin Institute of Technology, he joined University College Dublin, Ireland, in 2015, where he is currently an Associate Professor, the Director of the SFI Centre for Research Training in Machine Learning, and a Funded Investigator with the Insight and VistaMilk SFI research centres. He has co-authored the textbook Fundamentals of Machine Learning for Predictive Data Analytics: Algorithms, Worked Examples and Case Studies which was published in 2015 with MIT Press. His research interests include machine learning, and in particular novelty detection, human-in-the-loop machine learning, data visualisation for machine learning, and machine learning applications. 
\end{IEEEbiographynophoto}





\end{document}